\documentclass[runningheads]{llncs}
\usepackage{graphicx}
\usepackage{tabularx}
\usepackage{hyperref}
\usepackage{color}
\usepackage{amsmath}
\usepackage{amssymb}
\begin{document}
\title{L2AE-D: Learning to Aggregate Embeddings for Few-shot Learning with Meta-level Dropout}
%
%
\author{Heda Song\inst{1} \and
Mercedes Torres Torres\inst{2} \and
Ender \"Ozcan\inst{1} \and
Isaac Triguero\inst{1}
}
\authorrunning{H. Song et al.}
%

\institute{
	Automated Scheduling Optimisation and Planning (ASAP) Research Group
	\and
	Computer Vision Laboratory\\	
	University of Nottingham, Nottingham NG8 1BB, UK
	\email{\{heda.song,mercedes.torrestorres,ender.ozcan,isaac.triguero\}@nottingham.ac.uk}
}
\maketitle              
\begin{abstract}
Few-shot learning focuses on learning a new visual concept with very limited labelled examples. A successful approach to tackle this problem is to compare the similarity between examples in a learned metric space based on convolutional neural networks. However, existing methods typically suffer from meta-level overfitting due to the limited amount of training tasks and do not normally consider the importance of the convolutional features of different examples within the same channel. To address these limitations, we make the following two contributions: (a) We propose a novel meta-learning approach for aggregating useful convolutional features and suppressing noisy ones based on a channel-wise attention mechanism to improve class representations. The proposed model does not require fine-tuning and can be trained in an end-to-end manner. The main novelty lies in incorporating a shared weight generation module that learns to assign different weights to the feature maps of different examples within the same channel. (b) We also introduce a simple meta-level dropout technique that reduces meta-level overfitting in several few-shot learning approaches. In our experiments, we find that this simple technique significantly improves the performance of the proposed method as well as various state-of-the-art meta-learning algorithms. Applying our method to few-shot image recognition using Omniglot and miniImageNet datasets shows that it is capable of delivering a state-of-the-art classification performance.
\end{abstract}
\section{Introduction}
In recent years, deep learning techniques have dramatically been developed achieving high classification accuracy on visual recognition systems~\cite{he2016deep,hu2018squeeze}. These techniques usually require a large amount of labelled data to learn an appropriate model while they struggle when provided with very few data. However, in many real-world visual recognition tasks, such as images of new species or rare diseases, it is impractical to collect much labelled data. This highly restricts the successful application of deep learning. In addition, their learning style is typically not consistent with a human visual system that can generalise a new visual concept after seeing a few images based on previous experience. To address these issues, the computer vision community has raised enthusiasm for the challenge of learning from very few data, also known as few-shot learning~\cite{fei2006one,lake2011one}.

Few-shot learning typically aims to learn a new visual concept from a limited number of labelled examples. Overfitting can easily occur using conventional machine learning algorithms in such few-shot regime. To avoid this, we need a learning approach with a high generalisation ability. Inspired by the way humans are capable of quickly learning based on accumulated experience, many meta-learning approaches for few-shot learning have recently been proposed. In general, these methods learn a meta-learner to extract meta-knowledge from a distribution of few-shot learning tasks and further use it to assist unseen tasks. The extracted meta-knowledge can be represented by different algorithm components, such as a general feature extractor~\cite{koch2015siamese,vinyals2016matching,snell2017prototypical}, a distance metric~\cite{sung2018learning}, promising initial model parameters~\cite{finn2017model,alex2018reptile,finn2018probabilistic,lee2018gradient}, optimisation strategies~\cite{ravi2016optimization}, a model parameter predictor~\cite{Tsendsuren2017meta,Act2Param,gidaris2018dynamic}, a example generator~\cite{edwards2016towards,zhang2018metagan,gao2018low}, scale and/or shift vectors for activation adaptation~\cite{oreshkin2018tadam}, or label propagation~\cite{mishra2018simple,garcia2017few,liu2018learning}.

\begin{figure}[t!]
    \centering
	\includegraphics[width=1.0\linewidth]{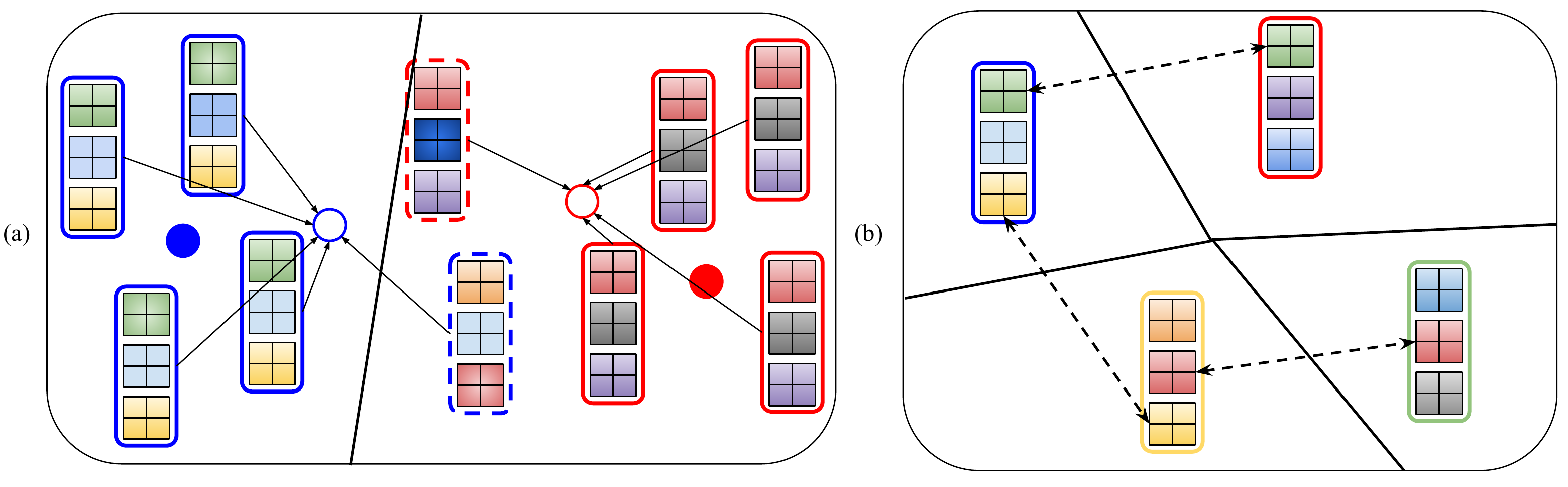}
	\caption{Illustration of our motivation. Each embedding (rounded rectangle) consists of three feature maps (coloured squares), with outliers shown in dashed borders. (a) Binary classification with five training examples per class. We show the real class centres in the embedding space (solid circles) and the mean of each class' embeddings (hollow circle). (b) 4-class classification  with one training example per class. Dashed arrows link similar feature maps in the embeddings from different classes.}
	\label{fig:motivation_1}
\end{figure}

Although these approaches achieve a good performance, they still suffer from several issues. Some methods need to fine-tune the base model when executing target tasks~\cite{finn2017model,alex2018reptile,finn2018probabilistic,lee2018gradient,ravi2016optimization}. Others introduce complex model architectures or external memory, which require more computing resources~\cite{garcia2017few,mishra2018simple,liu2018learning,Tsendsuren2017meta,gidaris2018dynamic}. Generative model-based approaches learn to generate more artificial examples, but they may create some non-informative examples when provided with noisy training examples~\cite{edwards2016towards,zhang2018metagan,gao2018low}. Metric learning based approaches are straightforward and efficient~\cite{snell2017prototypical}. They use Convolutional Neural Networks (CNNs) to extract the embeddings of examples, which are represented by a set of feature maps, and make predictions by comparing the similarities between embeddings. However, they seldom consider outliers in a class or borrow useful features from other classes~\cite{koch2015siamese,vinyals2016matching,snell2017prototypical,sung2018learning}. Due to the limited amount of data in few-shot learning, as presented in Fig.~\ref{fig:motivation_1}, the training examples may inherently contain uncertainties, such as the outliers shown in Fig.~\ref{fig:motivation_1}(a). If we simply use the mean of each class's embeddings as the class representative, as performed in~\cite{snell2017prototypical}, the possible outliers may force the representative to deviate from the class centre in the embedding space. Therefore, it is necessary to appropriately handle the effect of outliers. However, an outlier may actually contain some useful features, which could help strengthen part of the class representative. Similarly, even the examples of different classes may share some similar features as shown in Fig.~\ref{fig:motivation_1}(b), which could be used to help them to be distinguished from other classes in multi-class classification. Therefore, our goal is to reduce the impact of outliers and use as much as useful information as possible in few-shot learning.

In addition, these meta-learners may also suffer from overfitting. Although, meta-learners are trained on different few-shot learning tasks, they may consist of overlapped classes, because there are limited number of classes in the meta-training dataset and some of them are similar. For example, there are only 100 classes of objects in miniImageNet~\cite{ravi2016optimization} and some of them are different breeds of dogs. Thus, meta-learners could be trained to perform well on meta-training tasks and not generalise well on meta-testing tasks comprised of unseen classes.

To tackle the above issues, we propose L2AE-D (Learning to Aggregate Embeddings with Meta-level Dropout), a novel meta-learning approach for few-shot learning that learns to aggregate embeddings with meta-level dropout. L2AE-D learns a CNNs based feature extractor and a channel-wise attention mechanism in an end-to-end manner. The feature extractor is used to transform the input images into discriminative embeddings. The channel-wise attention mechanism is learned to assign larger weights to useful feature maps and smaller weights to noisy ones of different embeddings within the same channel. We propose different learning strategies for one-shot and few-shot tasks aiming to effectively exploit the few training embeddings. We also introduce a meta-level dropout technique into the meta-training process to prevent meta-level overfitting. We test this technique in several representative meta-learning approaches and it significantly improves their performance. We evaluate the proposed method on Omniglot~\cite{lake2011one} and miniImageNet~\cite{ravi2016optimization} datasets, and it achieves state-of-the-art performance.

\section{Related Work and Motivation}
Our method falls into the research field of few-shot learning. Section~\ref{fewshot-review} investigates the recent progress in this field, and explains our motivations. Besides, L2AE-D is based on an attention mechanism and dropout. We briefly review these techniques used in image classification in Section~\ref{attention-review} and~\ref{dropout-review}, respectively.

\subsection{Few-shot Learning approaches}
\label{fewshot-review}
Most recent works tackle few-shot learning by meta-learning due to its high generalisation ability. In general, they learn a meta-learner to extract meta-knowledge from a number of few-shot learning tasks and use it to assist in unseen ones. Depending on the type of meta-knowledge, these methods can be broadly classified into three categories. 

\noindent\textbf{Fast parametrisation based approaches:} Approaches in this class aim to learn a fast parametrisation strategy for quickly fine-tuning the base learner to adapt to new few-shot learning tasks. The most representative method, Model-Agnostic Meta-Learning (MAML)~\cite{finn2017model}, learns the model's initial parameters that can be adapted to task-specific model parameters by a few gradient descent steps based on few examples.
MAML has been extended in various ways, such as introducing a first-order gradient to reduce the computational burden~\cite{alex2018reptile}, learning model's initial parameters together with optimisation strategies (Meta-learner-LSTM~\cite{ravi2016optimization}), to further accelerate the fine-tuning process, choosing a subset of model parameters to fine-tune in order to make the model more task-specific~\cite{lee2018gradient}, modelling a distribution of prior model parameters to handle the inherent uncertainty of few-shot learning~\cite{finn2018probabilistic}. Rather than fine-tune the base model, some other methods learn a meta-learner to directly predict the parameters of the base model~\cite{Tsendsuren2017meta,gidaris2018dynamic,Act2Param}. One of them learns to predict the parameters of the fully-connected layer from the activations (Activation2Weights)~\cite{Act2Param}. These methods can be faster while some of them use external memory, which require more resources to store the adequate historical information. Instead, our approach executes target few-shot tasks in a feed-forward manner without external memory, which is quick and does not require additional resources.

\noindent\textbf{Generative model based approaches:} These approaches learn to generate artificial examples to compensate the lack of training data. The Neural Statistician approach learns to produce statistics of a dataset, such as mean or variance, which are used to specify a Gaussian distribution for generating data~\cite{edwards2016towards}. Other methods introduce generative adversarial networks to learn sharper decision boundaries (MetaGAN)~\cite{zhang2018metagan} or model the latent distribution of novel classes~\cite{gao2018low}. These meta-learners generate fake examples to assist few-shot learning tasks. However, these examples could be non-informative when the few training examples are not representative. Conversely, our method learns to aggregate useful information and suppress noisy information, which can be more stable. 

\noindent\textbf{Metric learning approaches:} The approaches in this class learn to compare the similarity between examples in a learned metric space. Most approaches learn a general feature extractor, which is usually represented by CNNs~\cite{koch2015siamese,vinyals2016matching,snell2017prototypical,sung2018learning,garcia2017few,mishra2018simple,liu2018learning}, to transform examples into embeddings and then compute the similarity between each pair of training and query embeddings based on weighted L1 distance (Siamese Nets~\cite{koch2015siamese}), cosine distance (Matching Nets~\cite{vinyals2016matching}), Euclidean distance (Prototypical Networks (ProtoNets)~\cite{snell2017prototypical}) or a learned distance metric (Relation Network (RN)~\cite{sung2018learning}). Finally, the queries can be classified by a linear~\cite{koch2015siamese,sung2018learning}, a k-nearest neighbours~\cite{snell2017prototypical} or a weighted k-nearest neighbours~\cite{vinyals2016matching} classifier. Some other approaches in this branch propagate label information from training examples to unlabelled query examples based on similarity~\cite{garcia2017few,mishra2018simple,liu2018learning}. Specifically, Transductive Propagation Network (TPN)~\cite{liu2018learning} and Graph Neural Networks (GNNs)~\cite{garcia2017few} learn a graph construction module and propagate labels within the graph. Another approach combines temporal convolutions and soft attention to propagate label information~\cite{mishra2018simple}. These methods also aggregate embeddings, but they treat each embedding as a whole. Instead, we aggregate feature maps in each channel, which could make use of more information, even from an outlier or an example from different classes.

\subsection{Attention Mechanisms}
\label{attention-review}
An attention mechanism aims to tell a machine learner where to focus, which is inspired by the human perception system. It has been extensively studied these years and applied to various machine learning tasks, such as machine translation~\cite{bahdanau2014neural} or image caption~\cite{xu2015show}. Recently, a few works have introduced attention mechanisms to CNNs for image classification tasks~\cite{jetley2018learn,woo2018cbam,hu2018squeeze}. Our method is related to~\cite{woo2018cbam,hu2018squeeze}. These two approaches both learn a channel attention module to assign different weights to different feature maps in each convolutional layer. Their aim is to emphasise useful features and suppress irrelevant ones for each example in large-scale image classification tasks. However, our goal is to handle uncertainty and fully use the few training examples in few-shot learning, so that we carry out an attention mechanism along a different dimension. Specifically, they learn a multi-layer perceptron to assign weights to the feature maps for a single embedding, while we learn CNNs to assign weights to the feature maps of different embeddings in the same channel. Besides, our attention module is shared between channels of all layers while they learn an attention module for each layer. It is noteworthy that several meta-learning approaches also introduce the attention mechanism to tackle few-shot learning problems~\cite{vinyals2016matching,mishra2018simple,gidaris2018dynamic}. However, we use it in different ways and for different purposes. The approaches in~\cite{vinyals2016matching,mishra2018simple} use attention to propagate labels based on the similarities between a query and training examples. The method in~\cite{gidaris2018dynamic} use attention to generate a classifier's weights for unseen classes. In contrast, our attention mechanism is used to assign different weights to the feature maps of different examples, aiming at handling uncertainty and fully using the few training examples.

\subsection{Dropout}
\label{dropout-review}
Dropout is a simple way to prevent neural networks from overfitting~\cite{srivastava2014dropout}. However, it is seldom applied to convolutional layers in CNNs, because the shared-filter architecture dramatically reduces the number of model parameters which reduce the model's capacity to overfit~\cite{srivastava2014dropout}. Still, the experimental results in~\cite{srivastava2014dropout} show performing dropout in convolutional layers can prevent overfitting and further improve the performance on image recognition tasks. Different from applying dropout on common machine learning tasks, we perform dropout in the meta-level to tackle meta-level overfitting problems, in which the dropped model is used for both training and testing examples during meta-training.

\section{L2AE-D: Learning to Aggregate Embeddings with Meta-level Dropout}
In this section, we describe the proposed learning to aggregate embeddings with meta-level dropout (L2AE-D) method. We define the problem of few-shot learning in Section~\ref{problem-setup}. Then, Section~\ref{model} describes our model consisting of an embedding, attention and distance module. The specific model architecture is discussed in Section~\ref{model-architecture}. Finally, Section~\ref{method-dropout} presents how we perform meta-level dropout.

\subsection{Problem Set-Up}
~\label{problem-setup}
Few-shot classification problems~\cite{fei2006one} aim to classify testing examples into one of $C$ unique classes based on $K$ labelled training examples for each of $C$ class, which is called $C$-way $K$-shot classification. For each $C$-way $K$-shot classification task, the training set $D_{_{train}}=\left \{ \left ( x_{i},y_{i} \right ) \right \}_{i=1}^{K\times C}$ contains $K\times C$ training examples and the testing set $D_{_{test}}$ contains $n$ testing examples that share the same label space with $D_{_{train}}$. In conventional machine learning, we could train a learner to predict the label for each testing example in $D_{_{test}}$ based on $D_{_{train}}$. However, the learner cannot be trained effectively based on such few training examples. 

A number of approaches including our method tackle the problem by meta-learning. Typically, we have three meta-sets, meta-training set $\boldsymbol{\mathfrak{D}}_{meta-train}$, meta-validation set $\boldsymbol{\mathfrak{D}}_{meta-validation}$ and meta-testing set $\boldsymbol{\mathfrak{D}}_{meta-test}$. Their respective label space is disjoint from each other. The meta-training set $\boldsymbol{\mathfrak{D}}_{meta-train}$ is used for training a meta-learner that generalises well across a distribution of few shot learning tasks, which is represented as $\boldsymbol{\mathfrak{D}}_{meta-train}=\left \{ \left ( D_{train}^{j},D_{test}^{j} \right ) \right \}_{j=1}^{N}$. The $\boldsymbol{\mathfrak{D}}_{meta-validation}$ set is used to select suitable hyper-parameters of the meta-learner. We can evaluate the meta-learner on the  $\boldsymbol{\mathfrak{D}}_{meta-test}$ set.

Since the $\boldsymbol{\mathfrak{D}}_{meta-train}$ set includes a large amount of different few shot classification tasks, it is best to train the meta-learner in an episode-based manner as proposed in~\cite{vinyals2016matching}. In each meta-training iteration, a single few shot classification ($\left ( D_{train}^{j},D_{test}^{j} \right ), j\in [1,N]$) is sampled to train the meta-learner based on its performance on $D_{test}^{j}$. We can also introduce the strategy of batch meta-training as done in~\cite{finn2017model}. Thus, in each meta-training iteration, we sample a batch of few shot classification to train the meta-learner.

\subsection{Model}
~\label{model}
\begin{figure}
	\begin{center}
		\includegraphics[width=0.9\linewidth]{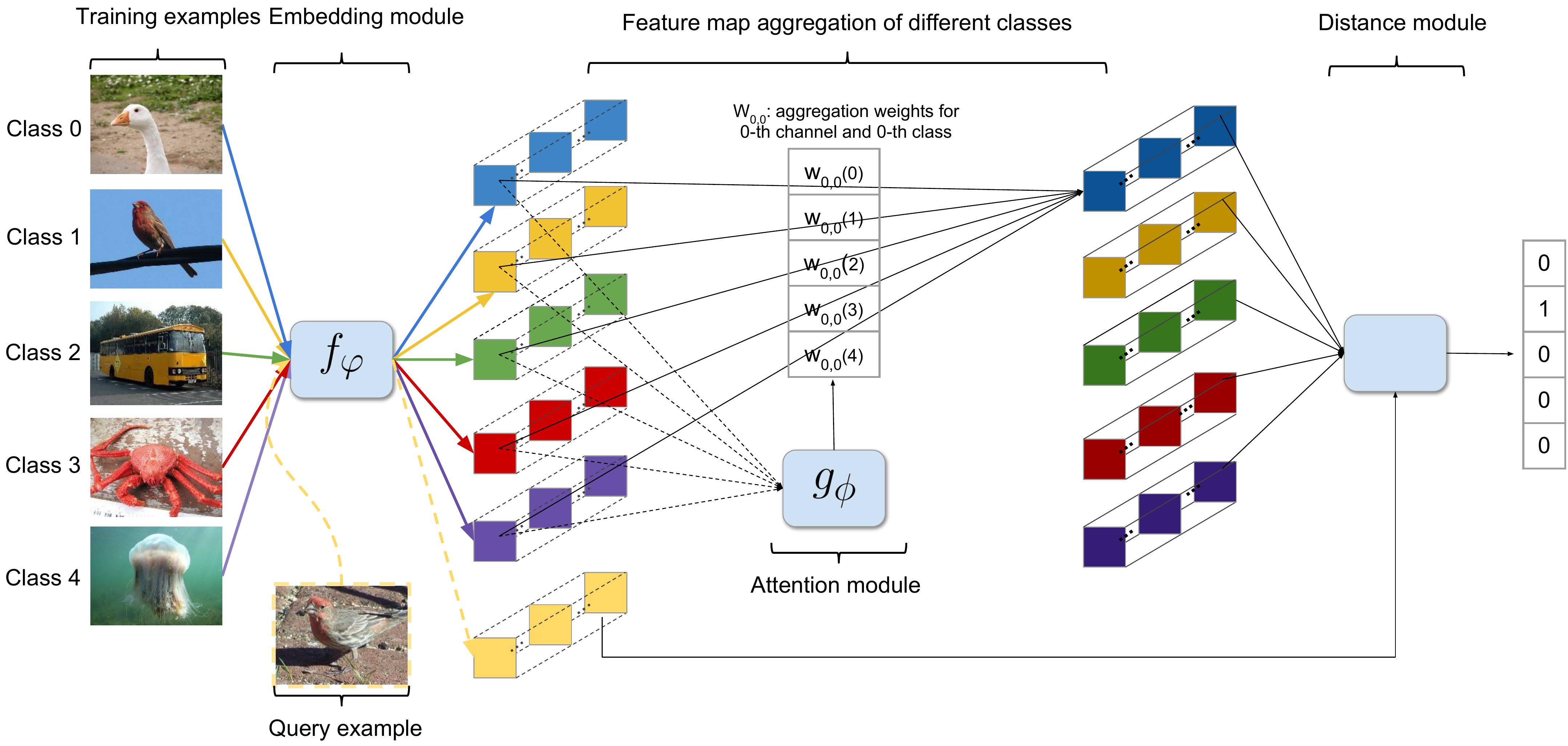}
	\end{center}
	\caption{5-way 1-shot classification with L2AE-D: (1) Training samples are transformed by $f_{\varphi }$ into embeddings (set of feature maps shown in coloured squares); (2) To strengthen the first feature map for the first class, we put it in the first channel and the other feature maps in the others, then we feed the concatenated 5-channel feature maps into $g_{\phi }$ to generate aggregation weights; (3) The 5 feature maps are aggregated based on the generated weights; (4) To make predictions, we feed a query into $f_{\varphi }$, then compare its embedding with the aggregated training embeddings in the distance module. This outputs a one-hot vector representing the predicted label of the query.}
	\label{fig:model_1}
\end{figure}

\begin{figure}
	\begin{center}
		\includegraphics[width=0.9\linewidth]{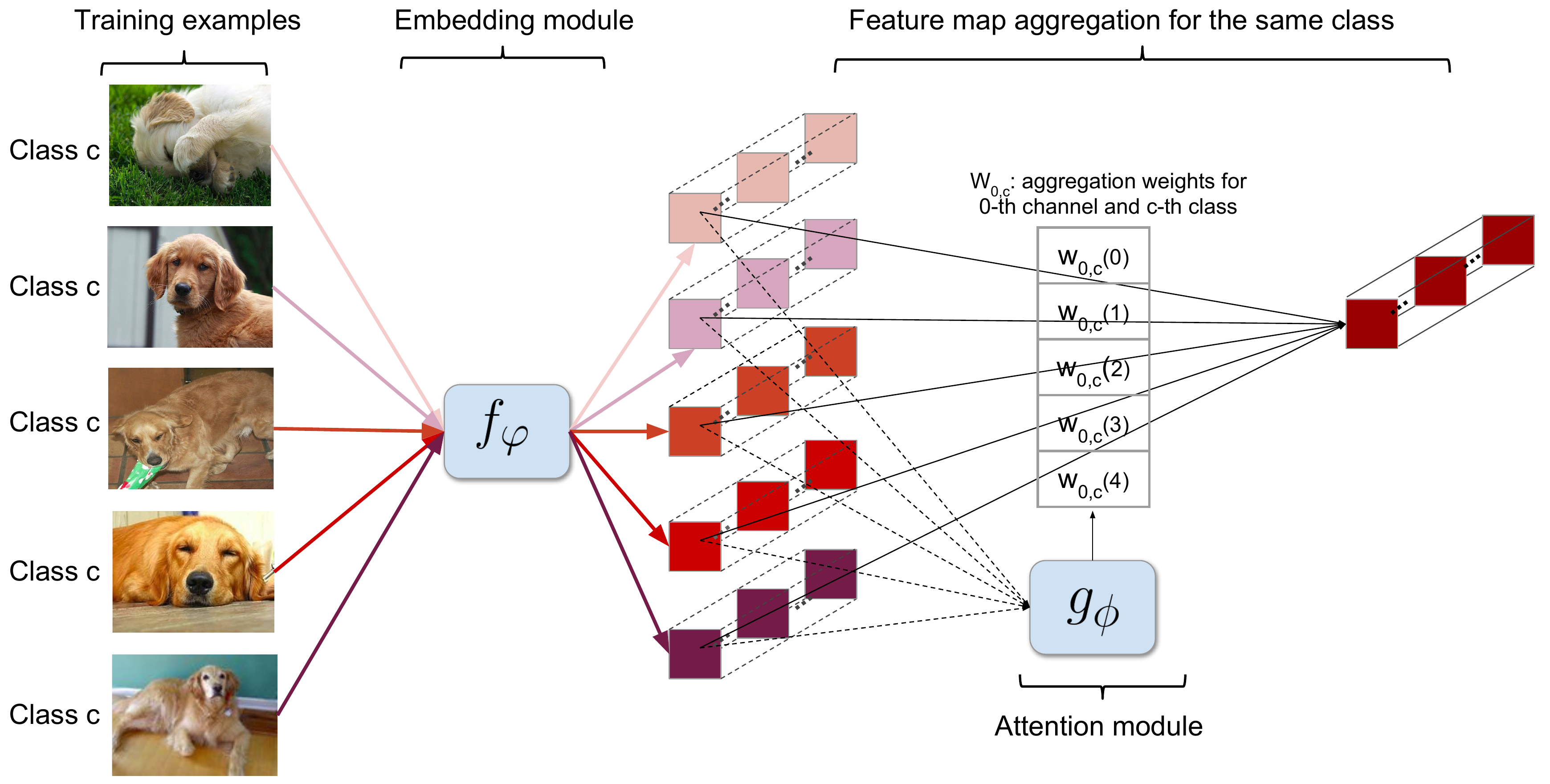}
	\end{center}
	\caption{C-way 5-shot classification with our approach. L2AE-D aggregates embeddings for each class: (1) The training examples are transformed by $f_{\varphi }$ into embeddings represented by a set of feature maps; (2) For each channel, we collect the feature maps and feed them into the attention module; (3) The feature maps are concatenated in depth and fed into $g_{\phi }$ to generate aggregation weights; (4) The feature maps are then aggregated based on the generated weights to represent a feature for this class.}
	\label{fig:model_2}
\end{figure}

L2AE-D can be divided into three modules: embedding module $f_{\varphi }$, attention module $g_{\phi }$ and distance module as shown in Fig. \ref{fig:model_1} and \ref{fig:model_2}. The attention module is different for the $1$-shot and $K$-shot cases. Fig.~\ref{fig:model_1} shows our strategy for $C$-way $1$-shot classification and Fig.~\ref{fig:model_2} depicts our strategy for $C$-way $K$-shot classification.

\noindent\textbf{Embedding module:} This module aims to extract features of each input image and transform it into embeddings. For each input example $x_{i}$ belonging to the $c$-th class, we feed it into the embedding module $f_{\varphi }$ to generate an embedding $E_{i,c}=\left \{ e^{1}_{i,c},e^{2}_{i,c},...,e^{n}_{i,c} \right \}$, which comprises $n$ feature maps $e^{k}_{i,c}\in R^{l\times l}$ with the size of $l\times l$. Then, the training embeddings are fed into the attention module.

\begin{figure}
	\begin{center}
		\includegraphics[width=0.9\linewidth]{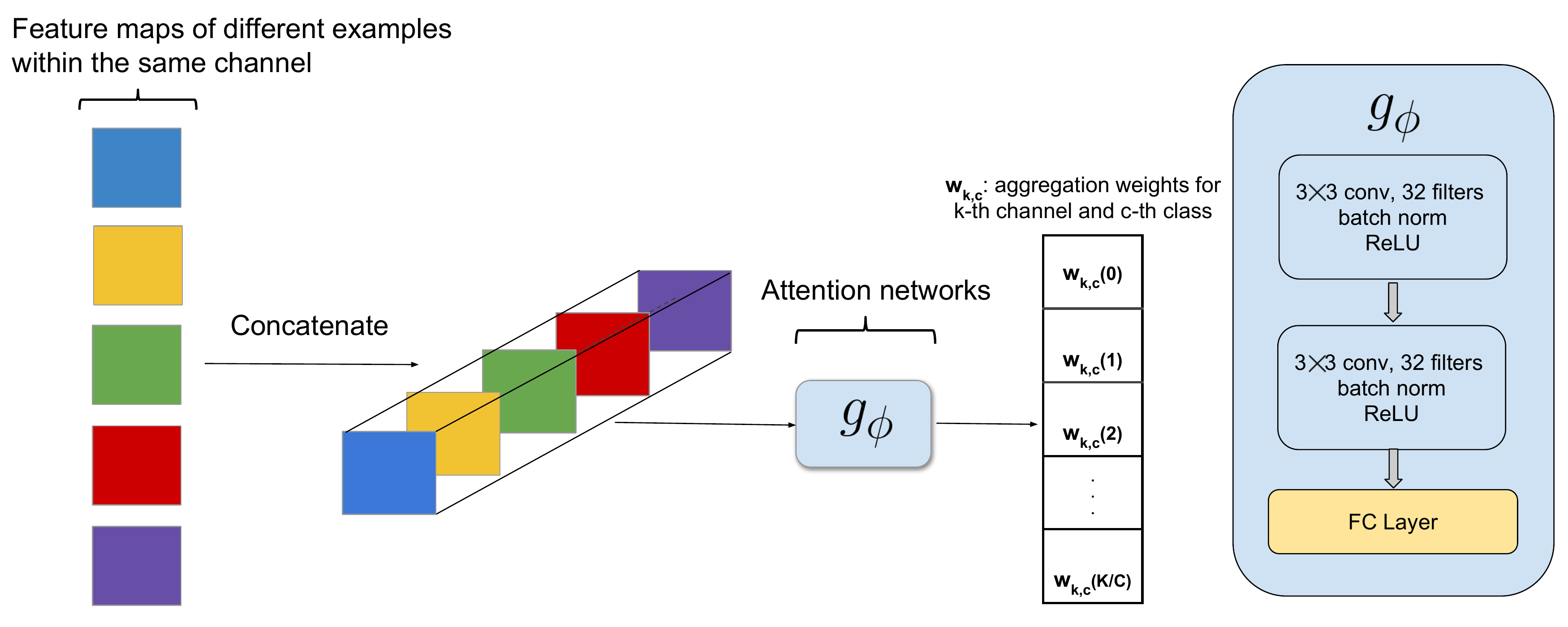}
	\end{center}
	\caption{The architecture of the attention module.}
	\label{fig:model_3}
\end{figure}

\noindent\textbf{Attention module:} This module is used for generating aggregation weights of the feature maps in a channel-wise manner as shown in Fig. \ref{fig:model_3}. Also, it is shared among different channels. We use two different strategies to do aggregation for $1$-shot and $K$-shot tasks as shown in Fig. \ref{fig:model_1} and \ref{fig:model_2}, respectively. For $K$-shot $C$-way tasks, we aggregate the feature maps of $K$ training embeddings in the same class to be the class-representative feature maps. For the $k$-th channel, we join the corresponding feature maps of $K$ training embeddings in the $c$-th class as $F_{k,c} = \left \{e^{k}_{1,c},e^{k}_{2,c},...,e^{k}_{K,c}\right \}$. For $C$-way $1$-shot tasks, we aggregate the feature maps of $C$ training embeddings from different classes, since there is only one training embedding in each class. To generate aggregation weights for the $c$-th class in the $k$-th channel, we concatenate the corresponding feature maps of $C$ training embeddings from different classes as $F_{k,c} = \left \{e^{k}_{1,c},e^{k}_{1,1},...,e^{k}_{1,C}\right \}$. Note that we locate $e^{k}_{1,c}$ in the first channel and other $C-1$ feature maps behind randomly in $F_{k,c}$. 

Next, the concatenated feature maps are inputted into CNNs based attention networks $g_{\phi }$, which produce the aggregation weights $w_{k,c}\in R^{K}$ for $K$-shot tasks or $w_{k,c}\in R^{C}$ for $1$-shot tasks. After that, we can aggregate the feature maps $F_{k,c}$ based on the weights $w_{k,c}$. The aggregated feature map of the $c$-th class in the $k$-th channel is represented by $\widetilde{e}^{k}_{c} = w_{k,c} \cdot F_{k,c} \in R^{l\times l}$. In the end, we concatenate the aggregated feature maps in all channels and obtain a new embedding $\widetilde{E}_{c}=\left \{ e^{1}_{c},e^{2}_{c},...,e^{n}_{c} \right \}$ for the $c$-th class. The new training embedding set is then represented by $\widetilde{E}_{train}=\left \{ \widetilde{E}_{c} \right \}^{C}_{c=1}$, in which $\widetilde{E}_{c}$ can be seen as a class representative.

\noindent\textbf{Distance module:} This module is used to measure the distance between the embeddings of query examples, $E_{q} = f_{\varphi }\left ( x_{q} \right )$, and the aggregated embeddings $\widetilde{E}_{train}$. Following ~\cite{snell2017prototypical}, we choose the Euclidean distance as the distance function $d(): R^{M}\times R^{M}\rightarrow [0,+\infty)$. Thus, the distance between $E_{q}$ and $\widetilde E_{c}$ is computed by $d(E_{q}, \widetilde E_{c})$.

\noindent\textbf{Loss function:} We consider cross-entropy loss to train our model. First, the softmax function is applied over the negative distance between the query embeddings and aggregated training embeddings as follows:
\begin{displaymath}
p_{\varphi,\phi }(y=c \mid x_{q})  = \frac{exp(-d(E_{q},\widetilde{E}_{c}))}{\sum_{c{}'} exp(-d(E_{q},\widetilde{E}_{c{}'}))}
\end{displaymath}
Then the loss function can be formulated as
\begin{displaymath}
L(\varphi,\phi )= -\sum_{q=1}^{N_{q}}\sum_{c=1}^{C} log\, p_{\varphi,\phi }(y=c \mid x_{q})
\end{displaymath}
where $N_{q}$ is the number of query examples in each training epoch.

\subsection{Model Architecture}
~\label{model-architecture}
L2AE-D follows the same architecture as the embedding module in prior approaches~\cite{vinyals2016matching,snell2017prototypical}, which contain 4 convolutional blocks. Each block is composed of a $3\times 3$ convolution with 64 filters, followed by batch normalisation (BN)~\cite{ioffe2015batch}, a ReLU nonlinearity and a $2\times 2$ max-pooling. For Omniglot, due to the small size of the input images, we feed the embeddings into the CNNs based attention module and we remove the max-pooling layer from the last convolutional block. The architecture of our attention module showed in Fig. \ref{fig:model_3} consists of 2 convolutional blocks and a fully connected (FC) layer. Each convolutional block in this module comprises a $3\times 3$ convolution with 32 filters, followed by batch normalisation, a ReLU nonlinearity. The FC layer results in a $m$-dimensional output, which represent the aggregation weights for the $m$ feature maps. For $K$-shot tasks, we use the softmax function after the FC layer since we aim to assign positive weights, whose sum is 1, to the $m$ embeddings of the same class.

\subsection{Meta-level Dropout}
~\label{method-dropout}
Most meta-learning approaches use multi-layer CNNs to extract features on few-shot learning. As discussed before, we incorporate the dropout technique in the meta-level to tackle meta-level overfitting.
Specifically, we randomly drop part of units of CNNs for each few-shot learning task during meta-training and the dropped model is used to extract features on both training and testing examples for each specific task. During meta-testing, we use the whole trained CNNs to extract features on both training and testing examples. Note that the dropout in the convolutional layers works in a different way from that in the fully connected layers, because the kernel weights are shared with the units at different spatial positions, so that, the weights would still be updated by backpropagation even if part of units are dropped. The actual effect of performing dropout in the convolutional layers is to scale the learning rate, which can also help with preventing overfitting. We find this technique can improve several meta-learning approaches significantly according to the experimental results in Section~\ref{drop-exp}.

\section{Experiments}
This section evaluates our method on the widely studied datasets Omniglot~\cite{lake2011one} and miniImageNet~\cite{ravi2016optimization}. The experimental setup is provided in Section~\ref{experiment-setting}. Section~\ref{result-omniglot} and~\ref{result-image} analyse the results on Omniglot and miniImageNet, respectively. We also introduce a meta-level dropout technique into several promising few-shot learning approaches and we test its behaviour on miniImageNet in Section~\ref{drop-exp}. Section \ref{visualise} visualises the working of L2AE-D based on T-distributed Stochastic Neighbor Embedding (t-SNE)~\cite{maaten2008visualizing}. The code for L2AE-D is available online\footnote{\url{github.com/Heda-Song/L2AE-D}}.

\subsection{Experimental setup}
\label{experiment-setting}
This section introduces the details of the two used datasets and the configurations followed to test the behaviour of L2AE-D against the state-of-the-art.

\begin{itemize}
    \item  \textbf{Omniglot} consists of 1,623 handwritten characters collected from 50 alphabets. There are 20 examples of each character, which are drawn by different people. We augmented the datasets with rotations with multiple 90 degrees as proposed by~\cite{santoro2016meta} to get 6492 classes. Following~\cite{finn2017model}, we randomly select 1,200 classes (4,800 classes after augmentation) for meta-training, 100 classes (400 classes after augmentation) for meta-validation, and the remaining 323 (1292 classes after augmentation) for meta-testing. All the input images are resized to $28 \times 28$ as suggested by~\cite{vinyals2016matching} to get a suitable sized embedding.

    \item \textbf{miniImageNet} was proposed by~\cite{vinyals2016matching} derived from the original ILSVRC-12 dataset~\cite{russakovsky2015imagenet}. It comprises 100 classes of colour images with 600 of each (60,000 in total). In our experiments, we use the widely used splits proposed by~\cite{ravi2016optimization}, which divides the 100 classes into 64 for meta-training, 16 for meta-validation and 20 for meta-testing. All the input images are resized to $84 \times 84$ as done by most few-shot learning approaches~\cite{ravi2016optimization,snell2017prototypical,finn2017model}. Note that the existing approaches use different tools to resize the images in miniImageNet. We use the library provided by OPENCV~\cite{opencv_library} following~\cite{liu2018learning}.

\end{itemize}

To allow for fair comparisons with the current state-of-the-art, we maintain the different experimental setups reported on Omniglot (20-way 5-shot, 20-way 1-shot, 5-way 5-shot, 5-way 1-shot) and miniImageNet (5-way 5-shot, 5-way 1-shot). All experiments are performed using TensorFlow~\cite{tensorflow} on a Titan V GPU.

\begin{itemize}
    \item \textbf{Meta-training:} Following most existing methods~\cite{snell2017prototypical,finn2017model,sung2018learning}, we train our model in an episode-based manner and use a meta-batch size of 4, which means in each episode we randomly sample 4 $C$-way $K$-shot classification tasks to train the model. For each few-shot task, besides the $C \times K$ training examples, we randomly sample 5 or 15 query examples per class to compute the loss for Omniglot and miniImageNet, respectively. We train our model with Adam~\cite{kingma2014adam} with a initial learning rate of $0.001$ in an end-to-end manner~\cite{snell2017prototypical,sung2018learning}. We cut the learning rate in half every 20,000 episodes to stabilise training and use meta-validation set to choose the best-performing model for meta-testing. It is noteworthy that existing methods conduct BN in different ways. As pointed out in~\cite{ravi2016optimization}, there would be a bad impact on performance if we use the global BN statistics accumulated from meta-training set to normalise batches of examples in meta-testing set, since there is no overlapping between the classes in these two sets. Thus, we perform BN on each batch of examples following~\cite{finn2017model,sung2018learning}. Specifically, for each task during both meta-training and meta-testing, we use each batch's statistics to normalise the training or query examples, which can be seen as a transductive way.
    
    \item \textbf{Meta-testing:} To be consistent with the existing few-shot learning approaches, we evaluate our model on 1,000 or 600 randomly sampled $C$-way $K$-shot classification tasks, which consist of $C \times K$ training examples and 5 or 15 query examples per class, for Omniglot and miniImageNet, respectively. We report the average accuracy on these tasks with 95\% confidence intervals. However, we find that most previous methods only use a single seed to randomly sample a batch of testing tasks and report the average accuracy. Since there are a large number of tasks in meta-testing, they may sample a large proportion of easy-to-classify or difficult-to-classify tasks using different seeds, which would lead to a result with high variance. To get a more reliable result, we use 10 different seeds to randomly sample different batches of testing tasks for 10 times and report the best, worst and average accuracy. Note that the existing methods are not strictly comparable since their experimental settings are not consistent with each other.
\end{itemize}

\subsection{Analysis of the Results on Omniglot}
\label{result-omniglot}
We compare our approach against state-of-the-art methods from each family of few-shot learning approaches that provide experimental results on Omniglot. They are MAML~\cite{finn2017model} from fast-parametrisation based approaches, Neural Statistician~\cite{edwards2016towards} and MetaGAN~\cite{zhang2018metagan} from generative model based approaches, and Siamese Nets~\cite{koch2015siamese}, Matching Nets~\cite{vinyals2016matching}, ProtoNets~\cite{snell2017prototypical}, GNN~\cite{garcia2017few} and RN~\cite{sung2018learning} as metric learning approaches. Their reported experimental results and ours are shown in Table \ref{tab:omni}. In general, all the methods perform worse on 20-way tasks than 5-way tasks, which shows 20-way tasks are more difficult. L2AE-D achieves state-of-the-art performance on 20-way tasks even in the worst case and competitive results on 5-way tasks. Besides, our results are very stable, since the differences between the best and worst accuracies for all the tasks are no more than 0.2\%. On 5-way 5-shot and 20-way 1-shot tasks, L2AE-D mostly obtains the best performance on different batches of tasks (using different seeds) since the average and best accuracy are the same. MetaGAN performs better on 5-way tasks by generating more examples to assist RN while it improves marginally upon RN.

\begin{table*}
    \caption{\small Few shot classification results on Omniglot averaged over 1,000 testing tasks. The $\pm$ shows 95\% confidence over tasks. The best-performing results are highlighted in bold. All the results are rounded to 1 decimal place other than MetaGAN's that are reported with 2 decimal places.}\label{tab:omni}
	\centering
	\begin{tabular}{lcccc}\hline
	    \bf Model & \multicolumn{2}{c}{\textbf{5-way Acc.}} & \multicolumn{2}{c}{\textbf{20-way Acc.}} \\
	    & 1-shot & 5-shot & 1-shot & 5-shot \\
		\hline
		\textbf{Siamese Nets} \cite{koch2015siamese}  &96.7\% &98.4\% &88.0\% &96.5\% \\ 
		\textbf{Matching Nets} \cite{vinyals2016matching} & 98.1\% & 98.9\% &93.8\% & 98.5\% \\
		\textbf{Neural Statistician} \cite{edwards2016towards} & 98.1\%& 99.5\%& 93.2\% &  98.1\%\\ 
		\textbf{ProtoNets} \cite{snell2017prototypical}  &98.8\% &99.7\%  &96.0\% &98.9\% \\ 
		\textbf{GNN} \cite{garcia2017few} &99.2 \% & 99.7\% &97.4\% &99.0 \% \\
		\textbf{MAML} \cite{finn2017model} &98.7$\pm$0.4\% & 99.9$\pm$0.1\% &95.8$\pm$0.3\% &98.9$\pm$0.2\% \\ 
		\textbf{RN} \cite{sung2018learning} & 99.6$\pm$0.2\% &99.8$\pm$0.1\% &97.6$\pm$0.2\% &99.1$\pm$0.1\%\\ 
		\textbf{MetaGAN} \cite{zhang2018metagan}+\textbf{RN} \cite{sung2018learning} & \textbf{99.67$\pm$0.18\%} &\textbf{99.86$\pm$0.11\%} &97.64$\pm$0.17\% &\textbf{99.21$\pm$0.1\%}\\
		\hline
		\textbf{L2AE-D (worst)} & 99.2$\pm$0.2\% &99.7$\pm$0.1\% &\textbf{97.7$\pm$0.2\%} &\textbf{99.2$\pm$0.1\%}\\ 
		\textbf{L2AE-D (average)} & 99.3$\pm$0.2\% &99.8$\pm$0.1\% &\textbf{97.8$\pm$0.2\%} &\textbf{99.2$\pm$0.1\%}\\
		\textbf{L2AE-D (best)} & 99.4$\pm$0.2\% &99.8$\pm$0.1\% &\textbf{97.8$\pm$0.2\%} &\textbf{99.3$\pm$0.1\%}\\
		\hline
	\end{tabular}
	\vspace{-1em}
\end{table*}

\subsection{Analysis of the Results on miniImageNet}
\label{result-image}
The existing few-shot learning approaches typically use two types of models to extract features, 4-layer CNNs~\cite{finn2017model,snell2017prototypical,sung2018learning} and deep residual networks~\cite{mishra2018simple,gao2018low,oreshkin2018tadam}. Deep residual network~\cite{he2016deep} is a kind of neural network with skip connections and more hidden layers, which has a more complex architecture but better representation capability compared to 4-layer CNNs. For a fair comparison, we compare our method with prior approaches that are based on the same type of model, 4-layer CNNs. As before, we choose state-of-the-art methods from each family that provide experimental results on miniImageNet. They are Meta-learner-LSTM~\cite{ravi2016optimization}, MAML~\cite{finn2017model} and Activation2Weights~\cite{Act2Param} from fast-parameterisation based approaches, MetaGAN~\cite{zhang2018metagan} from generative model based approaches, Matching Nets~\cite{vinyals2016matching}, ProtoNets~\cite{snell2017prototypical}, GNN~\cite{garcia2017few}, RN~\cite{sung2018learning} and TPN~\cite{liu2018learning} from metric learning approaches. Their reported experimental results and ours are shown in Table \ref{tab:imagenet}. L2AE-D achieves state-of-the-art performance on 5-way 5-shot classification even in the worst case. On 5-way 1-shot classification, L2AE-D (average) provides the second best result, which is slightly worse than Activations2Weights. However, the feature extractor of Activations2Weights is trained with more classes (higher ways) and more queries in each meta-training episode. In contrast, our model is trained on 5-way classification with 15 queries per episode, which is consistent with the setting of most existing approaches. Besides, TPN obtains very competitive results on 1-shot and 5-shot classification. However, TPN is a transductive method that requires unlabelled data to propagate labels and its performance is affected by the number of query examples. Even though we use query batch statistics to normalise the query examples in a transductive way, we can simply modify it into an inductive way by using training batch statistics to normalise the query data without decreasing the performance much.

\begin{table}
\caption{\small Few-shot classification results on miniImageNet averaged over 600 tests based on 4-layer CNNs. The $\pm$ shows 95\% confidence over tasks. FT stands for fine-tuning. The best-performing results are highlighted in bold.}
\centering
\begin{tabular}{lccc}\hline
            \\[-1em]
			\textbf{Model}             & \textbf{FT} & \multicolumn{2}{c}{\textbf{5-way Acc.}} \\
			&             & 1-shot             & 5-shot             \\ \hline
			\textbf{Matching Nets} \cite{vinyals2016matching} & N  & 43.56 $\pm$ 0.84\%       & 55.31 $\pm$ 0.73\%         \\
			\textbf{Meta-Learner-LSTM} \cite{ravi2016optimization} & N & 43.44 $\pm$ 0.77\%       & 60.60 $\pm$ 0.71\%       \\
			\textbf{MAML (1 query)} \cite{finn2017model}  & Y & 48.70 $\pm$ 1.84\% & 63.11 $\pm$ 0.92\%       \\
			\textbf{ProtoNets} \cite{snell2017prototypical} & N  & 49.42 $\pm$ 0.78\%       & 68.20 $\pm$ 0.66\%       \\
			\textbf{GNN} \cite{garcia2017few}  & N  & 50.33 $\pm$ 0.36\%       & 66.41 $\pm$ 0.63\%       \\
			\textbf{RN} \cite{sung2018learning} & N & 50.44 $\pm$ 0.82\%       & 65.32 $\pm$ 0.70\%  \\
			\textbf{MetaGAN}~\cite{zhang2018metagan}+\textbf{RN} \cite{sung2018learning} & N & 52.71 $\pm$ 0.64\%       & 68.63 $\pm$ 0.67\%  \\
			\textbf{TPN} \cite{liu2018learning}  & N  & 53.75 $\pm$ 0.86\%       & 69.43 $\pm$ 0.67\%       \\
			\textbf{Activations2Weights} \cite{Act2Param}  & N  & \textbf{54.53 $\pm$ 0.40\%}       & 67.87 $\pm$ 0.70\%       \\
			\hline
			\textbf{L2AE-D (worst)} & N    & 53.03 $\pm$ 0.84\%   & \textbf{69.53 $\pm$ 0.65\%}   \\
			\textbf{L2AE-D (average)} & N    & 53.85 $\pm$ 0.85\%   & \textbf{70.16 $\pm$ 0.65\%}   \\ 
			\textbf{L2AE-D (best)} & N    & 54.26 $\pm$ 0.87\%   & \textbf{70.76 $\pm$ 0.67\%}   \\
			\hline          
\end{tabular}
\label{tab:imagenet}
\end{table}

\subsection{Analysis of the effect of Meta-level Dropout}
\label{drop-exp}
Since the augmented Omniglot dataset includes much more classes (4,800) than miniImageNet (64) in the meta-training set, the meta-learners do not suffer much from meta-level overfitting on Omniglot. Therefore, we focus on miniImagent to analyse the effect of meta-level dropout through 5-way 1-shot tasks. We introduce meta-level dropout into several representative meta-learning approaches, including MAML~\cite{finn2017model}, ProtoNets~\cite{snell2017prototypical} and RN~\cite{sung2018learning}. Specifically, we use their provided code and add dropout in the middle two convolutional layers before max-pooling with the keep probability of 0.5, because there is more co-adaptation of features in the middle layers~\cite{yosinski2014transferable} and we find this setting can achieve better results. We compare the results with dropout to the reported ones of the chosen methods except MAML, since it tests on 1 query per class using 32 filters in CNNs. We modified their setting to use 64 filters and test on 15 queries per class in order to be consistent with the settings of other methods. We evaluate these methods on 5-way 1-shot classification in the same way as Section~\ref{result-image}. The experimental results in Table \ref{tab:drop} show that adding meta-level dropout can significantly improve several promising meta-learning approaches, as well as ours. It can also be seen that, even without dropout, L2AE also outperforms those representative few-shot learning approaches including ProtoNets that we improve upon.

\begin{table}
\caption{\small Few shot classification results on miniImageNet with or without dropout averaged over 600 testing tasks. The $\pm$ shows 95\% confidence over tasks.}
\centering
\begin{tabular}{lcc}\hline
			\\[-1em]
			\textbf{Model}        & \multicolumn{2}{c}{\textbf{5-way 1-shot Acc.}} \\
			& without dropout    & with dropout             \\ \hline
			\textbf{MAML (64filters, 15 queries)} \cite{finn2017model} & 47.71    $\pm$ 0.84\% & \textbf{50.43 $\pm$ 0.87\%}       \\
			\textbf{ProtoNets} \cite{snell2017prototypical} & 49.42 $\pm$ 0.78\%       & \textbf{52.08 $\pm$ 0.81\%}       \\
			\textbf{RN} \cite{sung2018learning} & 50.44 $\pm$ 0.82\%       & \textbf{52.40 $\pm$ 0.85\%}  \\ 
			\hline
			\textbf{L2AE}  & 51.55 $\pm$ 0.82\%   &  \textbf{53.85 $\pm$ 0.85\% (L2AE-D)}   \\
			\hline
			\\[-1em]
\end{tabular}
\label{tab:drop}
\end{table}

\subsection{Visualisation of the working of L2AE-D}
\label{visualise}
\begin{figure}
	\begin{center}
		\includegraphics[width=1.0\linewidth]{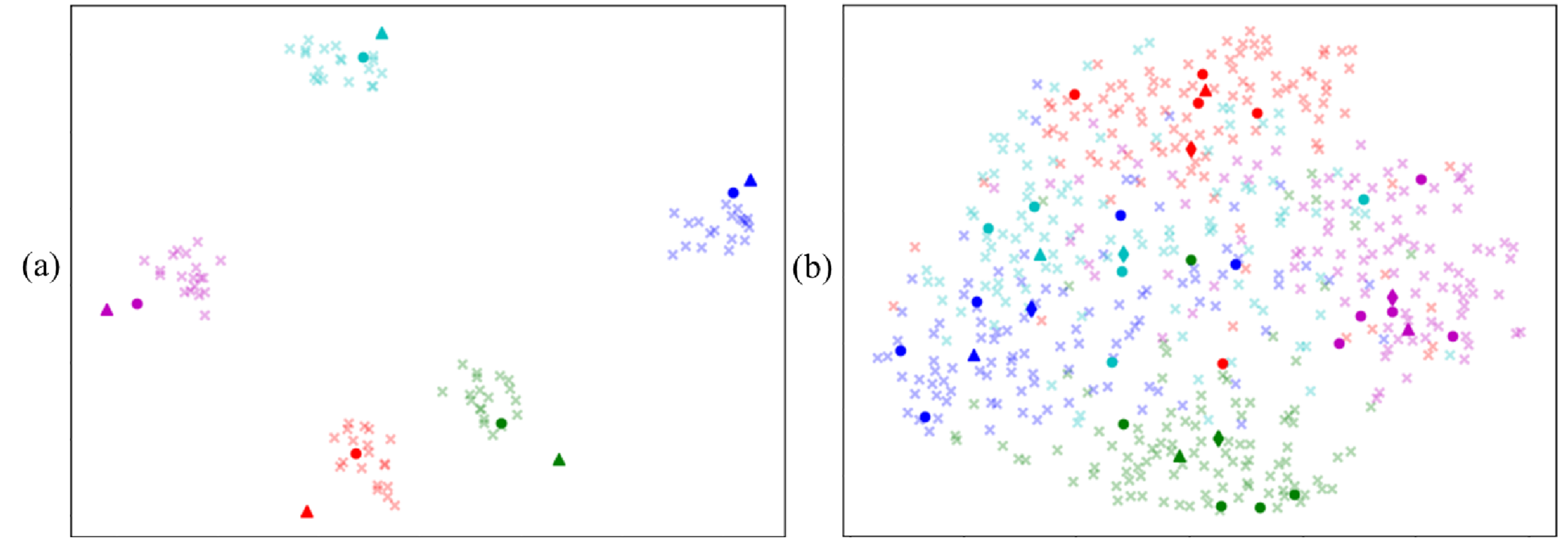}
	\end{center}
	\caption{t-SNE visualisation of the aggregated embeddings of unseen classes for a 5-way 1-shot classification task on Omniglot (a) and a 5-way 5-shot task on miniImagenet (b). The embeddings of training samples are shown as points. Aggregated embeddings are shown as triangles. The embeddings of regular examples are shown as crosses. The Means of training embeddings are shown as diamonds.}
	\label{fig:tsne-visualisation}
\end{figure}

To further show how our approach works, we visualise the aggregated embeddings for the unseen few-shot classification tasks in the meta-testing set based on t-SNE~\cite{maaten2008visualizing}. t-SNE is a technique for dimensionality reduction that is particularly well suited for the visualisation of high-dimensional datasets~\cite{maaten2008visualizing}. Fig.~\ref{fig:tsne-visualisation}(a) shows the visualisation of the aggregated embeddings for an unseen 5-way 1-shot classification task on Omniglot. The embeddings aggregated from different classes tend to move away from their own cluster and be farther from the clusters of other classes.

Fig.~\ref{fig:tsne-visualisation}(b) shows the visualisation of the aggregated embeddings for an unseen 5-way 5-shot classification task on miniImagenet. When there are unrepresentative examples in the training set, which means they are far from the cluster of their class, the mean of training embeddings~\cite{snell2017prototypical} deviates from a good position that represents a class in the embedding space. However, our aggregated embeddings stick to a representative position in the embedding space and are much more stable regardless of unrepresentative examples.

\section{Conclusions}
In this paper, we propose a novel meta-learning approach
for aggregating useful convolutional features and suppressing noisy ones based on a channel-wise attention mechanism. We propose two different learning strategies for one-shot and few-shot tasks aiming to fully and effectively use the few training examples. Our model does not require any fine-tuning and can be trained in an end-to-end manner. In addition, we tackle the problem of meta-level overfitting by introducing a meta-level dropout technique. This technique significantly improve several well-known meta-learning approaches as well as ours. Furthermore, we achieve state-of-the-art performance over 20-way classification tasks on Omniglot and 5-way tasks on miniImageNet, which demonstrate the effectiveness and competitiveness of our method.

\bibliographystyle{splncs04}
\bibliography{l2aed}

\end{document}